\pdfoutput=1

\documentclass[11pt]{article}

\usepackage[final]{acl}

\usepackage{times}
\usepackage{latexsym}
\usepackage{multirow}
\usepackage{booktabs}
\usepackage{array}
\usepackage[T1]{fontenc}

\usepackage[utf8]{inputenc}

\usepackage{microtype}

\usepackage{inconsolata}

\usepackage{graphicx}
\usepackage{algorithm}
\usepackage{algpseudocode}
\usepackage{amsmath}
\usepackage{enumitem}

\usepackage{xcolor}
\usepackage{pifont}  

\newcommand{\cmark}{\textcolor{green}{\ding{51}}} 
\newcommand{\xmark}{\textcolor{red}{\ding{55}}}   
\usepackage{latexsym}
\usepackage{listings}

\usepackage[normalem]{ulem}
\usepackage{microtype}
\usepackage{tikz}

\usepackage{subcaption}

\usepackage{xcolor}

\definecolor{codegreen}{rgb}{0,0.6,0}
\definecolor{codegray}{rgb}{0.5,0.5,0.5}
\definecolor{codepurple}{rgb}{0.58,0,0.82}
\definecolor{backcolour}{rgb}{0.95,0.95,0.92}

\lstdefinestyle{mystyle}{
    backgroundcolor=\color{backcolour},   
    commentstyle=\color{codegreen},
    keywordstyle=\color{magenta},
    numberstyle=\tiny\color{codegray},
    stringstyle=\color{codepurple},
    basicstyle=\ttfamily\footnotesize,
    breakatwhitespace=false,         
    breaklines=true,                 
    captionpos=b,                    
    keepspaces=true,                 
    showspaces=false,                
    showstringspaces=false,
    showtabs=false,                  
    tabsize=2
}

\title{LLM Agents Implement an NLG System from Scratch:
Building Interpretable Rule-Based RDF-to-Text Generators}

\author{Mateusz Lango \and Ond\v rej Du\v sek \\
 Charles University, Faculty of Mathematics and Physics, Prague, Czechia \\
   \texttt{\{lango,odusek\}@ufal.mff.cuni.cz} \\}

\begin{document}
\maketitle
\begin{abstract}
We present a novel neurosymbolic framework for RDF-to-text generation, in which the model is “trained” through collaborative interactions among multiple LLM agents rather than traditional backpropagation. The LLM agents produce rule-based Python code for a generator for the given domain, based on RDF triples only, with no in-domain human reference texts. The resulting system is fully interpretable, requires no supervised training data, and generates text nearly instantaneously using only a single CPU. 
Our experiments on the WebNLG and OpenDialKG data show that outputs produced by our approach reduce hallucination, with only slight fluency penalties compared to finetuned or prompted language models.
\end{abstract}

\begin{figure*}
    \centering
    \includegraphics[width=\textwidth]{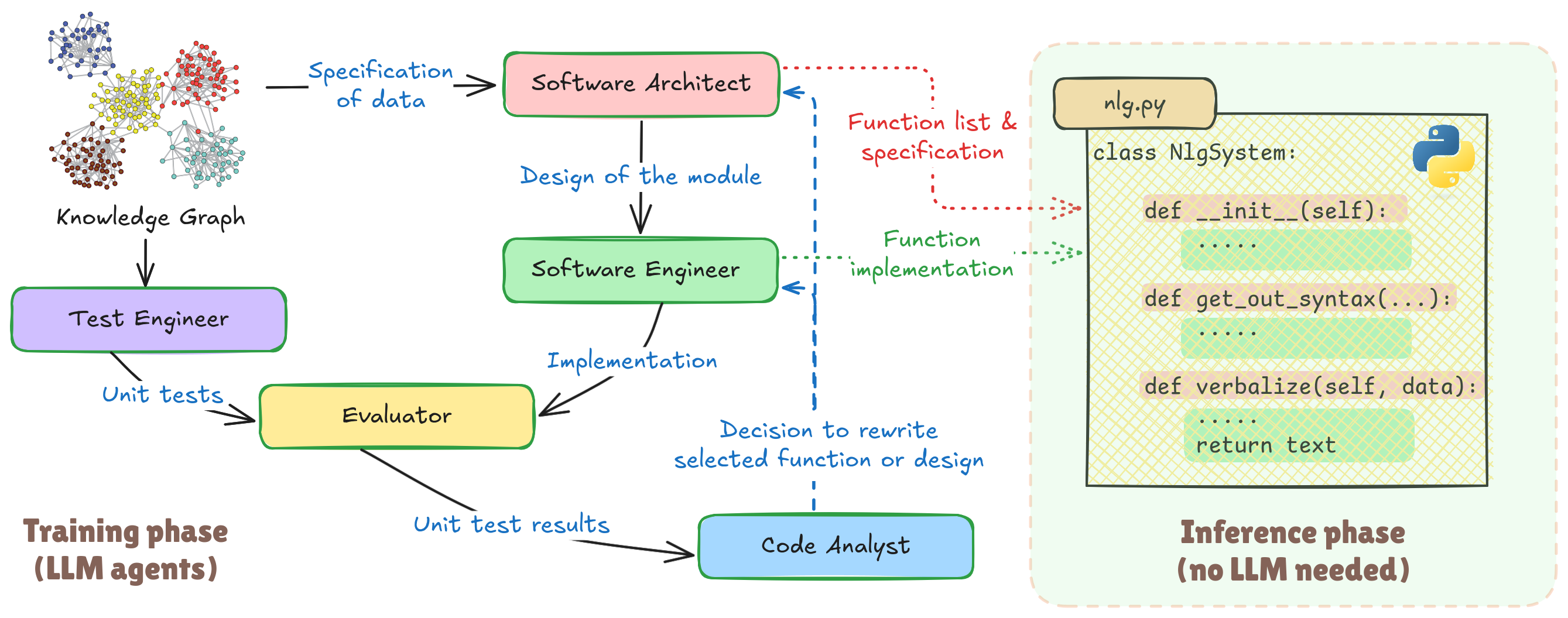}
    \caption{Overview of the presented approach. LLM Agents (boxes with green border) interact with each other to write an entire NLG system in pure Python during the training phase. The final system is fully interpretable, easy to edit by a human, and does not need any LLM during inference.}    
    \label{fig:overview}
\end{figure*}
\section{Introduction}
RDF-to-text is a popular task in natural language generation (NLG) that involves converting a subset of a knowledge graph, represented as RDF triples, into coherent natural language text~\cite{castro-ferreira-etal-2020-2020,agarwal-etal-2021-knowledge,kasner-dusek-2022-neural,10.1162/tacl_a_00641}. For instance, one possible verbalization of the following RDF triples: (Chopin, birthplace, Poland), (Chopin, birth year, 1810) is ``Chopin was born in 1810 in Poland.''

RDF-to-text systems are typically built using either rule-based or neural approaches~\cite{gattkrahmer}. Rule-based methods~\cite{lavoie-rainbow-1997-fast,white-baldridge-2003-adapting} use predefined templates and linguistic rules for precise, controlled output. In contrast, neural approaches rely on supervised learning from human data \cite{ke-etal-2021-jointgt,chen-etal-2020-kgpt} or in-context learning with large language models (LLMs) \cite{axelsson-skantze-2023-using,mille-etal-2024-2024} to generate more fluent and varied text,
yet their incorporation in industrial applications faces significant challenges.
Despite impressive benchmark performance, neural NLG systems generally lack interpretability and controllability, suffer from hallucinations, and require substantial computational resources 
 \cite{9521221,10.1145/3571730}.


In this work, we introduce a novel paradigm for building interpretable RDF-to-text systems that, instead of relying on supervised data, leverages the coding capabilities of large language models (LLM) to develop a full NLG system from scratch in pure Python.
Our approach involves a training stage where several LLM agents collaborate to iteratively design, implement, test, and refine a rule-based NLG model for a given domain using only unsupervised data (in-domain RDF triples, with no human references).
Once the training is complete, the system operates independently of any LLMs or neural components.

Experiments conducted on five datasets demonstrate that the proposed approach outperforms non-trivial neural baselines on reference-based metrics while offering full interpretability and controllability, producing fewer hallucinations, and providing remarkably fast inference times on a single CPU.

\section{Training a rule-based NLG system}
Our approach to training an NLG system relies on five LLM Agents. \emph{Software Architect} (SA) comes up with a design of the NLG system, making high-level decisions about the code structure. 
\emph{Software Engineer} (SE) iterates over the particular functions of the designed code structure and implements each one.
\emph{Evaluator} is a Python execution engine that runs the automatically written NLG system and then uses an LLM to assess the textual outputs produced. 
Unit tests for evaluation are supplied by \emph{Test Engineer}, embracing the test-driven development (TDD) paradigm for software development.
Finally, \emph{Code Analyst} (CA) analyses the NLG system implementation and any failing unit tests, determining whether the issues can be resolved by rewriting specific functions or if a full redesign of the system is needed. Depending on CA's decision, the training process returns to either SE or SA agent, which then revise the selected parts of NLG system accordingly. 
The approach is illustrated on Fig.~\ref{fig:overview} and in Appendix~\ref{app:code}.

The input to the training process is a knowledge graph, parts of which will later be verbalised by the constructed NLG system.
Note that no reference texts or annotated examples are used.
The output of the training is a single Python file containing the implementation of NLG system.
At inference time, the system is able to generalise to unseen data, provided it adheres to the same schema -- specifically, that predicates are defined consistently with those in the training graph.
We provide a more detailed description of each LLM agent involved in the training process below.

\paragraph{Test Engineer} begins by extracting a list of all predicates present in the knowledge graph (KG).
To provide the model with contextual understanding of each predicate, a random triple containing the predicate in question is selected from the graph. 
The LLM is then instructed to generate 50 input-output example pairs\footnote{
While our approach does not use generated pseudo-references during training, as the whole process is reference-less, we find that instructing the model to generate sets of input triples alongside pseudo-references results in more plausible examples.} for a data-to-text system using these predicates. Any examples containing predicates not found in KG are discarded, and the remaining examples are added to the set of unit tests. This process is repeated until each predicate is covered by at least three unit tests.
The exact prompts of TE and other agents are provided in~Appendix~\ref{app:prompt}.

\paragraph{Software Architect}
is given 
a list of all predicates found in the KG, along with an instruction to produce the high-level design of a rule-based NLG system.
SA's output defines the code structure by specifying a list of required functions, their responsibilities, input arguments, and interactions.
The only hardcoded requirement is the main entry point class and function.

\paragraph{Software Engineer}
iterates over the SA-produced list of functions and implements them one-by-one, given
a description of the design, the code implemented so far, and the signature of the function to be implemented. 
In the later stages of training, the SE is also given feedback from the Code Analyst and a list of failed unit tests.

\paragraph{Evaluator}
executes the NLG system code for each unit test within a Python interpreter, running each instance in a separate process with a predefined timeout, marking errors or timeouts as failures. 
Successful outputs are sent to an LLM, which answers a yes/no question on whether the generated verbalization correctly reflects the given input. To speed up evaluation, the process is terminated as soon as five failed unit tests are detected.
If the constructed program passes all unit tests, the training process is terminated.

\paragraph{Code Analyst}
receives the evaluation results and analyses both the system design and its current implementation to determine the root causes of the failed tests.  Based on this analysis, CA decides whether the issues stem from flaws in the overall design or from specific functions in the implementation.
If a full redesign is needed, the CA's textual feedback is passed back to SA, which produces a new design. 
If only certain functions require revision, CA supplies a list of these  functions to SE to reimplement. 

The interaction between the LLM agents, i.e.~the system training process, terminates either when the constructed NLG system passes all unit tests, or when the maximum iteration limit is reached.

\section{Experiments}
\subsection{Experimental setup}

\paragraph{Baselines}
We compare the results of our rule generation approach with two baselines: fine-tuned BART \citep[see Appendix~\ref{app:trainingdetails} for training details]{lewis-etal-2020-bart} and prompted Llama~3.3 70B \cite{touvron2024llama3} with a simple post-processing to remove superfluous text (see full prompt in Appendix~\ref{app:prompt}).

\paragraph{Datasets}
We experiment on two domains, with five datasets in total. 
First, the models were trained on the popular WebNLG domain~\cite{gardent-etal-2017-creating}, which contains data expressed as RDF triples alongside their corresponding text references. 
For evaluation, we used four test sets: the standard WebNLG test set and three datasets from the GEM 2024 shared task~\cite{mille-etal-2024-2024}. The GEM datasets were specifically designed to test system robustness by including RDF triples that are:
 (1) factual -- containing factually correct information;
 (2) counterfactual -- data from the factual dataset, with switched entity names;
 (3) fictional -- the triples contain fictional entities.

Second, we trained and evaluated the models on the OpenDialKG dataset~\cite{moon-etal-2019-opendialkg}, which contains dialogues annotated with RDF triples representing the information expressed in each utterance. We use this dataset for RDF-to-triple task, treating the utterances as textualisations of the data without taking dialogue history into account.

During training, our rule-based approach relied solely on the knowledge graph induced by the RDF triples from the dataset, but the fine-tuned neural baseline was trained using reference texts from the training set, with early stopping based on performance on the development set.

\paragraph{Our approach}
We tested our approach with three different LLMs: one proprietary LLM \citep[GPT-4.1][]{openai2025gpt41}) and three open-source models: Qwen 3 235B~\cite{qwen2024qwen25}, Qwen 2.5 72B~\cite{qwen2024qwen25} and Llama 3.3 70B~\cite{touvron2024llama3}.
The open-source models were used in 4-bit quantisation through the \texttt{ollama} library. Training was run with a maximum number of 25 iterations (10 for GPT) and repeated three times. The best model was selected based on the number of unit tests passed.
We use structured outputs to get an easy-to-process output from SA and CA.
As the entire WebNLG graph is substantial, we trained our system separately for each WebNLG thematic category.
As different LLMs are not equally strict when assessing the produced outputs, the Evaluator agent always used the Llama~3.3 model for better comparability.
The constructed programs are available in the code repository\footnote{\url{https://github.com/langus0/nlg-from-scratch}}.

\begin{table*}[t]
\small
\centering
\begin{tabular}{lc|rr|rr|rr|rr}
\toprule
                       & \bf Inter-                & \multicolumn{2}{c}{\bf BLEU}          & \multicolumn{2}{|c}{\bf METEOR}        & \multicolumn{2}{|c}{\bf BERTScore}     & \multicolumn{2}{|c}{\bf BLEURT} \\
                       & \bf pretability & \bf All             & \bf OOD             &\bf  All             & \bf OOD             & \bf All             & \bf OOD             & \bf All              & O\bf OD             \\\midrule
             
\multicolumn{9}{l}{\it Neural models}                                                                                                      \\\midrule
Fine-tuned BART        & \xmark                & \textbf{0.4352} & 0.3052          & 0.6791          & 0.6343          & \textbf{0.9308} & 0.9183          & 0.1275          & -0.0261         \\
Prompted Llama 3.3 70B & \xmark                & 0.3616          & 0.3327          & \uline{0.6887}          & \uline{0.6989}          & 0.9255          & 0.9243          & 0.1058          & 0.0969          \\\midrule
\multicolumn{9}{l}{\it Our rule-based NLG}       \\\midrule
trained by GPT-4.1                & \cmark              & { 0.3934}          & \uline{0.3615} & \textbf{0.7069} & \textbf{0.7124} & \uline{0.9291}          & \uline{0.9251} & \textbf{0.1841} & \uline{0.1483} \\
trained by Qwen 3 235B & \cmark & \uline{0.3939}&\bf 0.3772&0.6759&0.6980&0.9290&\bf 0.9281&\uline{0.1767}&\bf 0.1645\\
trained by Qwen 2.5 72B           & \cmark              & 0.3309          & 0.2609          & 0.6531          & 0.6456          & 0.9224          & 0.9175          &      0.1193           &     0.0655            \\
trained by Llama 3.3 70B          & \cmark         & 0.2858          & 0.2858          & 0.6578          & 0.6606          & 0.9179          & 0.9187          &      0.0762           &  0.0618               \\
\bottomrule        
\end{tabular}
\caption{Reference-based evaluation on the standard WebNLG test set. BLEU, METEOR, BERTScore and BLEURT metrics are reported for the entire test set (All) and for out-of-domain examples (OOD). }
\label{tab:mainr}
\end{table*}
\begin{table}[t]
\small
\centering
\begin{tabular}{lcccc}
\toprule

                       &\bf  BLEU          & \bf MET.           & \bf BERT.&\bf BLEURT              \\\midrule
\multicolumn{5}{l}{\it Neural models}                                                                                     \\\midrule
 BART        &\bf  0.9372&\bf 0.9849&\bf 0.9973&\bf 0.9340 \\
Llama 3.3 70b        & 0.2040&0.6289&0.9104&0.1348
   \\\midrule
\multicolumn{5}{l}{\it Our rule-based NLG}                                                                            \\\midrule
GPT 4.1        &0.3144&\uline{0.7313}&\uline{0.9272}&\uline{0.3247} \\
Qwen 3 235b        & \uline{0.3472}&0.6947&0.9239&0.1047  \\
Qwen 2.5 72b        & 0.3413&0.7030&0.9265&0.2300  \\
Llama 3.3 70B& 0.3120&0.6517&0.9216&0.1882 
\\\bottomrule     
\end{tabular}
\caption{Reference-based evaluation on the OpenDialKG dataset (MET.\,=\,METEOR, BERT.\,=\,BERTScore).}
\label{tab:mainr-open}
\end{table}

\subsection{Results of reference-based metrics}

We evaluate the quality of the generated outputs using several widely adopted reference-based metrics: BLEU~\cite{Papineni02bleu:a}, METEOR~\cite{meteor}, BERTScore~\cite{bert-score}, and BLEURT~\cite{bleurt}. 
This evaluation was not conducted on the GEM datasets, as they do not include reference texts.

The WebNLG test set results in Table~\ref{tab:mainr}   reveal that
our model trained by GPT-4.1 agents achieved the highest scores on METEOR and BLEURT metrics. Although a fine-tuned neural model outperformed ours on BLEU and BERTScore overall, our system still achieved better scores on these metrics within a more challenging subset of out-of-domain examples.
Our model also outperformed prompted Llama~3.3 70B. Note that neither of these systems was trained on human-written reference texts.

There is relatively little difference in performance between our rule-based systems produced by GPT 4.1 and those produced by the largest open-source model, Qwen 3 235B. While GPT 4.1 achieved better results on METEOR, BERTScore and BLEURT, Qwen 3 performed slightly better on BLEU and on out-of-domain examples.

NLG systems trained using smaller open-source LLMs were less successful, indicating that more powerful LLMs may be necessary for implementing complete NLG systems. Nonetheless, these models retain certain advantages over purely neural models, as they provide full transparency of the generation process and can potentially be manually improved by skilled developers.
The inference time comparision\footnote{The reported times do not include loading the models into memory and were measured on a machine with an Nvidia A40 48 GB GPU and an AMD EPYC 7313 CPU.} in Table~\ref{tab:time} shows another advantage of our models: they achieve a 35x speedup on CPU compared to the BART model running on GPU and 272x speedup while running both models on CPU.

\begin{table}[]
\small
\centering
\begin{tabular}{lrr}
\toprule
                       & \multicolumn{2}{c}{\bf Inference time} \\
                       &\bf GPU              &\bf CPU             \\\midrule
Fine-tuned BART        & 249 s            & 1910 s          \\
Prompted Llama 3.3 70B & 6360 s           & n/a             \\\midrule
Our approach (GPT-4.1)         & -                & \textbf{7 s}            \\\bottomrule
\end{tabular}
\caption{Inference time for the WebNLG test set.}
\label{tab:time}
\end{table}

The results obtained on OpenDialKG are presented in Table~\ref{tab:mainr-open}. Here, the fine-tuned model clearly obtained the highest results on reference-based metrics, indicating the importance of using the original training data to produce the expected sentence structures. Nevertheless, all of our models outperformed the prompted LLM on all metrics.

\begin{table*}[t]
\small
\setlength{\tabcolsep}{5pt}
\begin{tabular}{llll|lll|lll|lll}
\toprule
                       & \multicolumn{3}{c|}{\bf WebNLG test set}              & \multicolumn{3}{c|}{\bf GEM2 Counterfactual}          & \multicolumn{3}{c|}{\bf GEM2 Fictional}      & \multicolumn{3}{c}{\bf GEM2 Factual}                 \\
                       &\bf  Gram.          & \bf Add.           & \bf Om.           & \bf Gram.          & \bf Add.           &\bf  Om.           & Gram. & \bf Add.           & \bf Om.           & \bf Gram.          & \bf Add.           & \bf Om.           \\\midrule
\multicolumn{10}{l}{\it Neural models}                                                                                                        &                &                &                \\\midrule
 BART        & 0.692          & 0.510          & 0.526          & 0.426          & 0.613          & 0.622          & 0.619 & 0.580          & 0.599          & 0.689          & 0.527          & 0.512          \\
Llama 3.3  & \textbf{0.752} & 0.044          & {0.080} & \textbf{0.818} & 0.209          & \textbf{0.080}          & \textbf{0.937} & \textbf{0.018} & \uline{0.096 }         & \textbf{0.984} & {0.027} & {0.076}          \\\midrule
\multicolumn{10}{l}{\it Our rule-based NLG}                                                                                                &                &                &                \\\midrule
GPT-4.1                & \uline{0.734}          & {0.029} & 0.111          & \uline{0.517}          &  \uline{0.069} & {0.128} &\uline{0.738} &  {0.036} & {0.098} & \uline{0.730}          &  {0.034}           &0.108\\
Qwen 3 235B&0.729&\uline{0.026}&\uline{0.040}&0.392&0.071&0.106&0.632&0.043&\bf 0.066&\uline{0.730}&\uline{0.026}&\bf 0.040\\
Qwen 2.5 72B&0.663&\textbf{0.019}&\textbf{0.065}&0.440&\textbf{0.054}&\uline{0.091}&0.603&\uline{0.030} &{0.098}&0.660&\textbf{0.021}&\uline{0.066}\\
Llama 3.3 70B&0.635&0.049&0.126&0.419&0.070&0.138&0.551&0.065&0.208&0.633&0.050&0.127
\\\bottomrule     
\end{tabular}
\caption{Reference-less evaluation on four test sets: the standard WebNLG test set and three GEM 2024 shared task test sets. 
\uline{Gram}maticality, \uline{add}ition of unsupported facts, and \uline{om}issions are evaluated by an LLM-as-a-Judge.
}\label{tab:llm}
\end{table*}

\begin{table}[t]
\small
\centering
\begin{tabular}{llll}
\toprule

                       &\bf  Gram.          & \bf Add.           & \bf Om.              \\\midrule
\multicolumn{4}{l}{\it Neural models}                                                                                     \\\midrule
BART        &  0.502&0.052&0.139  \\
Llama 3.3        & \textbf{0.985}&0.030&0.063
   \\\midrule
\multicolumn{4}{l}{\it Our rule-based NLG}                                                                            \\\midrule
trained by GPT 4.1        & 0.923&\bf 0.013&\bf 0.022   \\
trained by Qwen 3 235B        & 0.840&0.018&0.106 \\
trained by Qwen 2.5 72B        & 0.666&0.033&0.146 \\
trained by Llama 3.3 70B& 0.598&0.056&0.075
\\\bottomrule     
\end{tabular}
\caption{Reference-less evaluation on the OpenDialKG test set (see Table~\ref{tab:llm} for metrics). 
}\label{tab:llm-open}
\end{table}

\subsection{Results of reference-less metrics}
We perform a reference-less evaluation on all test sets using the LLM-as-a-Judge approach~\cite{zheng2023judging,gu2025surveyllmasajudge}.
The selected LLM (Llama 3.3 70B) provides binary judgments on three aspects:  grammatical correctness of the generated text (Gram.), presence of unsupported facts (Add.), omission of input triples in the output (Om.). The exact prompts are provided in  Appendix~\ref{app:prompt-eval}.

 The results of systems trained on the WebNLG dataset shown in Table~\ref{tab:llm} reveal that the outputs of our rule-based system trained with GPT-4.1 are more grammatically correct and contain fewer hallucinations than the output of fine-tuned BART on all four test sets.
 The outputs produced by an LLM (Llama 3.3) achieve the highest grammatical correctness.
 On three out of four test sets, our model trained with Qwen~2.5 reduces the number of additions compared to the  LLM’s output -- sometimes by nearly fourfold -- while maintaining a comparable or lower number of omissions. 

The results for the OpenDialKG dataset in Table~\ref{tab:llm-open} show that our GPT-4.1-trained system produced significantly fewer additions and omissions (t-test, $\alpha=5\%$) than both fine-tuned BART and Llama 3.3.
Our model also achieved better grammatical correctness than BART while scoring slightly worse than Llama 3.3.

\subsection{Ablation experiments}
We performed two ablation experiments: 1) we replaced SA agent with a static system design  produced by a human (Abl.~1); 2) we used training examples from WebNLG training set instead of generated unit tests (Abl.~2).
The results of the 
ablations are in Table~\ref{tab:ablation}.
Using a static design of the system has a highly negative impact, which is especially visible in trainable metrics such as BLEURT. 
Evaluating using the original WebNLG training set examples instead of automatically generated unit tests also yields slightly worse results, demonstrating the utility of our approach.

\begin{table}[t]
\small
\setlength{\tabcolsep}{4pt}
\begin{tabular}{lccc>{\hspace{-1mm}}c}
\toprule
                       &\bf BLEU  &\bf MET. &\bf BERT. & \bf BLEURT \\
                       \midrule
Ours                   & 0.331 & 0.653  & 0.922     & 0.119  \\
Abl. 1 (design) & 0.323 & 0.611  & 0.912     &      -0.010\phantom{-}  \\
Abl. 2 (training set)  & 0.309 & 0.638  & 0.919     & 0.071 \\
\bottomrule
\end{tabular}
\caption{Ablation experiments on the WebNLG test set with Qwen 2.5 (see Table~\ref{tab:mainr} and~\ref{tab:mainr-open} for metrics).}
\label{tab:ablation}
\end{table}

\subsection{Human evaluation}
\label{sec:hum}
\begin{table}[t]
\small\centering
\begin{tabular}{l|ccccc}
\toprule
\bf  & \bf  min. h. & \bf maj. h.& \bf omi. & \bf disfl. & \bf rep.  \\
\midrule
BART &  0.22          & {0.40} & 0.25          & {0.20} & 0.08         \\
Llama 3  & {0.07}          & {0.05}         & \textbf{0.06} & 0.22          & \textbf{0.02}  \\

Our (GPT) & \textbf{0.00}          & \textbf{0.00}          & \textbf{0.06}          & \textbf{0.19}          & \textbf{0.02}                 \\
\bottomrule
\end{tabular}
\caption{Results of human evaluation: percentage of examples with \uline{min}or and \uline{maj}or \uline{h}allucinations, \uline{omi}ssions, \uline{disfl}uencies, \uline{rep}etitions.}
\label{tab:manual}
\end{table}
We conducted a small-scale in-house human evaluation for 100 randomly selected instances from the WebNLG test set. Outputs of our system (with GPT 4.1) and both baselines (BART, Llama 3.3) 
were annotated by six NLP experts who answered binary questions about the presence of minor hallucinations (e.g. typos in named entity names), major hallucinations (output containing facts not supported by the data), omissions (missing information), disfluencies (grammar errors or difficult-to-read text) and repetitions (information mentioned twice).
In total, 300 system outputs were annotated.
The interannotator agreement, measured by Cohen's Kappa and averaged over all questions, was 0.8288.

The results are presented in Table~\ref{tab:manual}. The annotators did not detect any hallucinations in the outputs of our system, indicating that our system generates very few hallucinations. Although our system occasionally omits facts from the input, its omission rate is comparable to that of a prompted LLM.
It also achieved the lowest assessment of disfluencies present in the generated text, and the smallest number of repetitions ex aequo with the prompted LLM.

\subsection{Evaluation of interpretability}
Since the result of training of our rule-based NLG approach is  Python code, it should be possible to understand how the text was produced and even modify it if needed.
We asked two experienced\footnote{One junior developer with two years of industrial experience, and one senior developer with 10+ years of experience.} Python software engineers (SEs) to get familiar with the implementation of our NLG system produced by GPT~4.1 and  perform two tasks:
\begin{itemize}
\item \emph{Interpretability task} -- we provided 25 examples of input triples and outputs produced by the system. In the output text, one word was randomly highlighted and the SEs were asked to provide the line number containing code that produced that word. If removing the indicated line from the code resulted in a text that did not contain the highlighted word, the test was considered as passed.
\item \emph{Modification task} -- we took all outputs of our system involving omissions, as indicated by human evaluators in Sec.~\ref{sec:hum}, and we asked SEs to modify the NLG system code to produce output without omissions. During this test, SEs could use an IDE of their choice, with the possibility of using a Python interpreter for testing, but no AI code assistants such as GitHub Copilot.  The outputs of the corrected systems were assessed by a human evaluator to estimate if the generated text still contains omissions.
\end{itemize}

All tests related to both tasks were successfully passed by the SEs. The average time taken to successfully complete the interpretability task for a single instance was 9.6 seconds. According to the SEs, the code was fully understandable, but it contained some unused parts and could be refactored to improve its clarity. The modification task required more time for code editing, but in almost all cases, this did not exceed five minutes. 
\subsection{How do the generated programmes look?}
On average, a program generated by our approach contains 168 lines of code. A typical NLG system groups RDF triples by subject, processes each group by adding modifiers to the subject, converts the group into a clause and then refines it into a sentence. To improve fluency, modifier ordering is often applied. 
Different LLMs exhibit varying coding styles, e.g.~Qwen 2.5 tends to produce Python code with typing. The generated code frequently imports standard Python libraries such as \texttt{datetime} or \texttt{defaultdict}, but occasionally also relies on less common ones like \texttt{inflect}, \texttt{num2words} or even \texttt{nltk}. While no runtime errors were observed when testing on the WebNLG dataset, evaluation on the GEM datasets produced some errors as the generated programs were not robust enough to handle differences in date formatting between the datasets. This resulted in reduced performance on these sets.
\section{Related work}

\paragraph{Program Synthesis} is the task of automatically generating programs from specifications, traditionally using formal methods \cite{gulwani2017program} or evolutionary search~\cite{koza1994genetic}, and increasingly leveraging neural networks~\cite{wyrwinski}. Modern approaches synthesize programs from natural language, input-output examples, and partial sketches. 

\paragraph{LLMs for Coding} Recently, Large Language Models trained on large corpora of code and natural language have exhibited remarkable code generation capabilities, enabling them to perform tasks such as code completion, code synthesis from natural language prompts, and bug fixing~\cite{chen2021evaluatinglargelanguagemodels, li2023starcodersourceyou}.
Beyond single-pass generation, reflective approaches like Reflexion \cite{shinn2023reflexion} and Self-Refine \cite{madaan2023selfrefineiterativerefinementselffeedback} introduced iterative frameworks that equip models with the ability to critique and revise their own outputs to improve constructed programs. These techniques are typically only employed to generate a single function for algorithmic tasks. Drawing inspiration from evolutionary program search, \citet{novikov2025alphaevolvecodingagentscientific} recently presented  AlphaEvolve framework, which uses an LLM ensemble to evolve more complex programs.
To the best of our knowledge, however, these approaches have not previously been applied to NLG system construction or more generally to the implementation of programs involving language processing.

\paragraph{LLMs for NLG template construction}
Recently, \citet{warczynski-etal-2024-leveraging-large} proposed a rule-based NLG systems that use LLM-written templates tailored to specific combinations of a triplet’s predicates. These systems rely on a hardcoded engine that splits input triples into known combinations, applies the corresponding templates, and merges the results into a single output text. Unlike our approach, this method requires a dataset with reference texts and does not generalize to out-of-domain examples. While technically interpretable, the method's interpretability is limited by the high number of templates it generates (over 113,000 for the WebNLG dataset) which also makes the produced systems difficult to maintain. We include a comparison with this approach in Appendix~\ref{app:rule-base}.

\section{Summary}
This paper presents a new approach to building RDF-to-text systems that uses neural LLMs to train a rule-based system written entirely in Python.
The resulting natural language generation (NLG) system is fully interpretable, enabling human intervention to modify its behaviour. The system generates text in a non-autoregressive manner, 
offering a significant improvement in speed over neural models.
Experimental results demonstrate that, although neural models excel at fluency, our approach is often competitive and reduces hallucinations.
\section*{Limitations}
Although the presented approach reduces the number of hallucinated texts, it may still generate non-factual outputs. The NLG system should undergo thorough testing before deployment.

\section*{Acknowledgments}

This work was supported by the European Research Council (Grant agreement No.~101039303, NG-NLG) and used resources of the LINDAT/\hspace{0mm}CLARIAH-CZ Research Infrastructure (Czech Ministry of Education, Youth, and Sports project No. LM2018101).

\bibliography{anthology,custom}

\begin{thebibliography}{35}
\providecommand{\natexlab}[1]{#1}

\bibitem[{Agarwal et~al.(2021)Agarwal, Ge, Shakeri, and
  Al-Rfou}]{agarwal-etal-2021-knowledge}
Oshin Agarwal, Heming Ge, Siamak Shakeri, and Rami Al-Rfou. 2021.
\newblock \href {https://doi.org/10.18653/v1/2021.naacl-main.278} {Knowledge
  graph based synthetic corpus generation for knowledge-enhanced language model
  pre-training}.
\newblock In \emph{Proceedings of the 2021 Conference of the North American
  Chapter of the Association for Computational Linguistics: Human Language
  Technologies}, pages 3554--3565, Online. Association for Computational
  Linguistics.

\bibitem[{Axelsson and Skantze(2023)}]{axelsson-skantze-2023-using}
Agnes Axelsson and Gabriel Skantze. 2023.
\newblock \href {https://aclanthology.org/2023.mmnlg-1.5/} {Using large
  language models for zero-shot natural language generation from knowledge
  graphs}.
\newblock In \emph{Proceedings of the Workshop on Multimodal, Multilingual
  Natural Language Generation and Multilingual WebNLG Challenge (MM-NLG 2023)},
  pages 39--54, Prague, Czech Republic. Association for Computational
  Linguistics.

\bibitem[{Banerjee and Lavie(2005)}]{meteor}
Satanjeev Banerjee and Alon Lavie. 2005.
\newblock \href {https://www.aclweb.org/anthology/W05-0909} {{METEOR}: An
  automatic metric for {MT} evaluation with improved correlation with human
  judgments}.
\newblock In \emph{Proceedings of the {ACL} Workshop on Intrinsic and Extrinsic
  Evaluation Measures for Machine Translation and/or Summarization}, pages
  65--72, Ann Arbor, Michigan. Association for Computational Linguistics.

\bibitem[{Castro~Ferreira et~al.(2020)Castro~Ferreira, Gardent, Ilinykh,
  van~der Lee, Mille, Moussallem, and
  Shimorina}]{castro-ferreira-etal-2020-2020}
Thiago Castro~Ferreira, Claire Gardent, Nikolai Ilinykh, Chris van~der Lee,
  Simon Mille, Diego Moussallem, and Anastasia Shimorina. 2020.
\newblock \href {https://aclanthology.org/2020.webnlg-1.7/} {The 2020
  bilingual, bi-directional {W}eb{NLG}+ shared task: Overview and evaluation
  results ({W}eb{NLG}+ 2020)}.
\newblock In \emph{Proceedings of the 3rd International Workshop on Natural
  Language Generation from the Semantic Web (WebNLG+)}, pages 55--76, Dublin,
  Ireland (Virtual). Association for Computational Linguistics.

\bibitem[{Chen et~al.(2021)Chen, Tworek, Jun, Yuan, de~Oliveira~Pinto, Kaplan,
  Edwards, Burda, Joseph, Brockman, Ray, Puri, Krueger, Petrov, Khlaaf, Sastry,
  Mishkin, Chan, Gray, Ryder, Pavlov, Power, Kaiser, Bavarian, Winter, Tillet,
  Such, Cummings, Plappert, Chantzis, Barnes, Herbert-Voss, Guss, Nichol,
  Paino, Tezak, Tang, Babuschkin, Balaji, Jain, Saunders, Hesse, Carr, Leike,
  Achiam, Misra, Morikawa, Radford, Knight, Brundage, Murati, Mayer, Welinder,
  McGrew, Amodei, McCandlish, Sutskever, and
  Zaremba}]{chen2021evaluatinglargelanguagemodels}
Mark Chen, Jerry Tworek, Heewoo Jun, Qiming Yuan, Henrique~Ponde
  de~Oliveira~Pinto, Jared Kaplan, Harri Edwards, Yuri Burda, Nicholas Joseph,
  Greg Brockman, Alex Ray, Raul Puri, Gretchen Krueger, Michael Petrov, Heidy
  Khlaaf, Girish Sastry, Pamela Mishkin, Brooke Chan, Scott Gray, and 39
  others. 2021.
\newblock \href {https://arxiv.org/abs/2107.03374} {Evaluating large language
  models trained on code}.
\newblock \emph{Preprint}, arXiv:2107.03374.

\bibitem[{Chen et~al.(2020)Chen, Su, Yan, and Wang}]{chen-etal-2020-kgpt}
Wenhu Chen, Yu~Su, Xifeng Yan, and William~Yang Wang. 2020.
\newblock \href {https://doi.org/10.18653/v1/2020.emnlp-main.697} {{KGPT}:
  Knowledge-grounded pre-training for data-to-text generation}.
\newblock In \emph{Proceedings of the 2020 Conference on Empirical Methods in
  Natural Language Processing (EMNLP)}, pages 8635--8648, Online. Association
  for Computational Linguistics.

\bibitem[{Gardent et~al.(2017)Gardent, Shimorina, Narayan, and
  Perez-Beltrachini}]{gardent-etal-2017-creating}
Claire Gardent, Anastasia Shimorina, Shashi Narayan, and Laura
  Perez-Beltrachini. 2017.
\newblock \href {https://doi.org/10.18653/v1/P17-1017} {Creating training
  corpora for {NLG} micro-planners}.
\newblock In \emph{Proceedings of the 55th Annual Meeting of the Association
  for Computational Linguistics (Volume 1: Long Papers)}, pages 179--188,
  Vancouver, Canada. Association for Computational Linguistics.

\bibitem[{Gatt and Krahmer(2018)}]{gattkrahmer}
Albert Gatt and Emiel Krahmer. 2018.
\newblock \href {https://arxiv.org/abs/1703.09902} {Survey of the state of the
  art in natural language generation: core tasks, applications and evaluation}.
\newblock \emph{J. Artif. Int. Res.}, 61(1):65–170.

\bibitem[{Gu et~al.(2025)Gu, Jiang, Shi, Tan, Zhai, Xu, Li, Shen, Ma, Liu,
  Wang, Zhang, Wang, Gao, Ni, and Guo}]{gu2025surveyllmasajudge}
Jiawei Gu, Xuhui Jiang, Zhichao Shi, Hexiang Tan, Xuehao Zhai, Chengjin Xu, Wei
  Li, Yinghan Shen, Shengjie Ma, Honghao Liu, Saizhuo Wang, Kun Zhang, Yuanzhuo
  Wang, Wen Gao, Lionel Ni, and Jian Guo. 2025.
\newblock \href {https://arxiv.org/abs/2411.15594} {A survey on
  llm-as-a-judge}.
\newblock \emph{Preprint}, arXiv:2411.15594.

\bibitem[{Gulwani et~al.(2017)Gulwani, Polozov, Singh
  et~al.}]{gulwani2017program}
Sumit Gulwani, Oleksandr Polozov, Rishabh Singh, and 1 others. 2017.
\newblock Program synthesis.
\newblock \emph{Foundations and Trends{\textregistered} in Programming
  Languages}, 4(1-2):1--119.

\bibitem[{Ji et~al.(2023)Ji, Lee, Frieske, Yu, Su, Xu, Ishii, Bang, Madotto,
  and Fung}]{10.1145/3571730}
Ziwei Ji, Nayeon Lee, Rita Frieske, Tiezheng Yu, Dan Su, Yan Xu, Etsuko Ishii,
  Ye~Jin Bang, Andrea Madotto, and Pascale Fung. 2023.
\newblock \href {https://doi.org/10.1145/3571730} {Survey of hallucination in
  natural language generation}.
\newblock \emph{ACM Comput. Surv.}, 55(12).

\bibitem[{Kasner and Dusek(2022)}]{kasner-dusek-2022-neural}
Zden{\v{e}}k Kasner and Ondrej Dusek. 2022.
\newblock \href {https://doi.org/10.18653/v1/2022.acl-long.271} {Neural
  pipeline for zero-shot data-to-text generation}.
\newblock In \emph{Proceedings of the 60th Annual Meeting of the Association
  for Computational Linguistics (Volume 1: Long Papers)}, pages 3914--3932,
  Dublin, Ireland. Association for Computational Linguistics.

\bibitem[{Ke et~al.(2021)Ke, Ji, Ran, Cui, Wang, Song, Zhu, and
  Huang}]{ke-etal-2021-jointgt}
Pei Ke, Haozhe Ji, Yu~Ran, Xin Cui, Liwei Wang, Linfeng Song, Xiaoyan Zhu, and
  Minlie Huang. 2021.
\newblock \href {https://doi.org/10.18653/v1/2021.findings-acl.223}
  {{J}oint{GT}: Graph-text joint representation learning for text generation
  from knowledge graphs}.
\newblock In \emph{Findings of the Association for Computational Linguistics:
  ACL-IJCNLP 2021}, pages 2526--2538, Online. Association for Computational
  Linguistics.

\bibitem[{Koza(1994)}]{koza1994genetic}
John~R Koza. 1994.
\newblock \emph{Genetic programming II: automatic discovery of reusable
  programs}.
\newblock MIT press.

\bibitem[{Lavoie and Rainbow(1997)}]{lavoie-rainbow-1997-fast}
Benoit Lavoie and Owen Rainbow. 1997.
\newblock \href {https://doi.org/10.3115/974557.974596} {A fast and portable
  realizer for text generation systems}.
\newblock In \emph{Fifth Conference on Applied Natural Language Processing},
  pages 265--268, Washington, DC, USA. Association for Computational
  Linguistics.

\bibitem[{Lewis et~al.(2020)Lewis, Liu, Goyal, Ghazvininejad, Mohamed, Levy,
  Stoyanov, and Zettlemoyer}]{lewis-etal-2020-bart}
Mike Lewis, Yinhan Liu, Naman Goyal, Marjan Ghazvininejad, Abdelrahman Mohamed,
  Omer Levy, Veselin Stoyanov, and Luke Zettlemoyer. 2020.
\newblock \href {https://doi.org/10.18653/v1/2020.acl-main.703} {{BART}:
  Denoising sequence-to-sequence pre-training for natural language generation,
  translation, and comprehension}.
\newblock In \emph{Proceedings of the 58th Annual Meeting of the Association
  for Computational Linguistics}, pages 7871--7880, Online. Association for
  Computational Linguistics.

\bibitem[{Li et~al.(2023)Li, Allal, Zi, Muennighoff, Kocetkov, Mou, Marone,
  Akiki, Li, Chim, Liu, Zheltonozhskii, Zhuo, Wang, Dehaene, Davaadorj,
  Lamy-Poirier, Monteiro, Shliazhko, Gontier, Meade, Zebaze, Yee, Umapathi,
  Zhu, Lipkin, Oblokulov, Wang, Murthy, Stillerman, Patel, Abulkhanov, Zocca,
  Dey, Zhang, Fahmy, Bhattacharyya, Yu, Singh, Luccioni, Villegas, Kunakov,
  Zhdanov, Romero, Lee, Timor, Ding, Schlesinger, Schoelkopf, Ebert, Dao,
  Mishra, Gu, Robinson, Anderson, Dolan-Gavitt, Contractor, Reddy, Fried,
  Bahdanau, Jernite, Ferrandis, Hughes, Wolf, Guha, von Werra, and
  de~Vries}]{li2023starcodersourceyou}
Raymond Li, Loubna~Ben Allal, Yangtian Zi, Niklas Muennighoff, Denis Kocetkov,
  Chenghao Mou, Marc Marone, Christopher Akiki, Jia Li, Jenny Chim, Qian Liu,
  Evgenii Zheltonozhskii, Terry~Yue Zhuo, Thomas Wang, Olivier Dehaene, Mishig
  Davaadorj, Joel Lamy-Poirier, João Monteiro, Oleh Shliazhko, and 48 others.
  2023.
\newblock \href {https://arxiv.org/abs/2305.06161} {Starcoder: may the source
  be with you!}
\newblock \emph{Preprint}, arXiv:2305.06161.

\bibitem[{Li et~al.(2024)Li, Li, Geng, Yang, Li, Yuan, He, Yuan, Ma, Huang, and
  Li}]{10.1162/tacl_a_00641}
Shujie Li, Liang Li, Ruiying Geng, Min Yang, Binhua Li, Guanghu Yuan, Wanwei
  He, Shao Yuan, Can Ma, Fei Huang, and Yongbin Li. 2024.
\newblock \href {https://doi.org/10.1162/tacl_a_00641} {Unifying structured
  data as graph for data-to-text pre-training}.
\newblock \emph{Transactions of the Association for Computational Linguistics},
  12:210--228.

\bibitem[{Madaan et~al.(2023)Madaan, Tandon, Gupta, Hallinan, Gao, Wiegreffe,
  Alon, Dziri, Prabhumoye, Yang, Gupta, Majumder, Hermann, Welleck,
  Yazdanbakhsh, and
  Clark}]{madaan2023selfrefineiterativerefinementselffeedback}
Aman Madaan, Niket Tandon, Prakhar Gupta, Skyler Hallinan, Luyu Gao, Sarah
  Wiegreffe, Uri Alon, Nouha Dziri, Shrimai Prabhumoye, Yiming Yang, Shashank
  Gupta, Bodhisattwa~Prasad Majumder, Katherine Hermann, Sean Welleck, Amir
  Yazdanbakhsh, and Peter Clark. 2023.
\newblock \href {https://arxiv.org/abs/2303.17651} {Self-refine: Iterative
  refinement with self-feedback}.
\newblock \emph{Preprint}, arXiv:2303.17651.

\bibitem[{Mille et~al.(2024)Mille, Sedoc, Liu, Clark, Axelsson, Clinciu, Hou,
  Mahamood, Obonyo, and Zhang}]{mille-etal-2024-2024}
Simon Mille, Jo{\~a}o Sedoc, Yixin Liu, Elizabeth Clark, Agnes~Johanna
  Axelsson, Miruna~Adriana Clinciu, Yufang Hou, Saad Mahamood, Ishmael~Nyunya
  Obonyo, and Lining Zhang. 2024.
\newblock \href {https://aclanthology.org/2024.inlg-genchal.2/} {The 2024 {GEM}
  shared task on multilingual data-to-text generation and summarization:
  Overview and preliminary results}.
\newblock In \emph{Proceedings of the 17th International Natural Language
  Generation Conference: Generation Challenges}, pages 17--38, Tokyo, Japan.
  Association for Computational Linguistics.

\bibitem[{Moon et~al.(2019)Moon, Shah, Kumar, and
  Subba}]{moon-etal-2019-opendialkg}
Seungwhan Moon, Pararth Shah, Anuj Kumar, and Rajen Subba. 2019.
\newblock \href {https://doi.org/10.18653/v1/P19-1081} {{O}pen{D}ial{KG}:
  Explainable conversational reasoning with attention-based walks over
  knowledge graphs}.
\newblock In \emph{Proceedings of the 57th Annual Meeting of the Association
  for Computational Linguistics}, pages 845--854, Florence, Italy. Association
  for Computational Linguistics.

\bibitem[{Novikov et~al.(2025)Novikov, Vũ, Eisenberger, Dupont, Huang, Wagner,
  Shirobokov, Kozlovskii, Ruiz, Mehrabian, Kumar, See, Chaudhuri, Holland,
  Davies, Nowozin, Kohli, and
  Balog}]{novikov2025alphaevolvecodingagentscientific}
Alexander Novikov, Ngân Vũ, Marvin Eisenberger, Emilien Dupont, Po-Sen Huang,
  Adam~Zsolt Wagner, Sergey Shirobokov, Borislav Kozlovskii, Francisco J.~R.
  Ruiz, Abbas Mehrabian, M.~Pawan Kumar, Abigail See, Swarat Chaudhuri, George
  Holland, Alex Davies, Sebastian Nowozin, Pushmeet Kohli, and Matej Balog.
  2025.
\newblock \href {https://arxiv.org/abs/2506.13131} {Alphaevolve: A coding agent
  for scientific and algorithmic discovery}.
\newblock \emph{Preprint}, arXiv:2506.13131.

\bibitem[{OpenAI(2025)}]{openai2025gpt41}
OpenAI. 2025.
\newblock \href {https://openai.com/index/gpt-4-1/} {Introducing gpt-4.1 in the
  api}.
\newblock Accessed: 2025-05-19.

\bibitem[{Papineni et~al.(2002)Papineni, Roukos, Ward, and jing
  Zhu}]{Papineni02bleu:a}
Kishore Papineni, Salim Roukos, Todd Ward, and Wei jing Zhu. 2002.
\newblock \href {https://www.aclweb.org/anthology/P02-1040} {{BLEU}: a method
  for automatic evaluation of machine translation}.
\newblock In \emph{Proceedings of the 40th annual meeting of the Association
  for Computational Linguistics}, pages 311--318, Philadelphia, PA, USA.

\bibitem[{Sellam et~al.(2020)Sellam, Das, and Parikh}]{bleurt}
Thibault Sellam, Dipanjan Das, and Ankur~P. Parikh. 2020.
\newblock \href {https://aclanthology.org/2020.acl-main.704/} {{BLEURT}:
  {Learning} {Robust} {Metrics} for {Text} {Generation}}.
\newblock In \emph{Proceedings of the 58th {Annual} {Meeting} of the
  {Association} for {Computational} {Linguistics}}, pages 7881--7892, Online.

\bibitem[{Shinn et~al.(2023)Shinn, Cassano, Berman, Gopinath, Narasimhan, and
  Yao}]{shinn2023reflexion}
Noah Shinn, Federico Cassano, Edward Berman, Ashwin Gopinath, Karthik
  Narasimhan, and Shunyu Yao. 2023.
\newblock \href {https://arxiv.org/abs/2303.11366} {Reflexion: Language agents
  with verbal reinforcement learning}.
\newblock \emph{Preprint}, arXiv:2303.11366.

\bibitem[{Touvron et~al.(2024)Touvron, Lavril, Izacard, Martinet, Lachaux,
  Lacroix, Rozière, Goyal, Simonyan, and Jegou}]{touvron2024llama3}
Hugo Touvron, Thibaut Lavril, Gautier Izacard, Xavier Martinet, Marie-Anne
  Lachaux, Timothée Lacroix, Baptiste Rozière, Naman Goyal, Karen Simonyan,
  and Hervé Jegou. 2024.
\newblock \href {https://arxiv.org/abs/2407.21783} {The llama 3 herd of
  models}.
\newblock \emph{arXiv preprint arXiv:2407.21783}.

\bibitem[{Warczy{\'n}ski et~al.(2024)Warczy{\'n}ski, Lango, and
  Dusek}]{warczynski-etal-2024-leveraging-large}
J{\k{e}}drzej Warczy{\'n}ski, Mateusz Lango, and Ondrej Dusek. 2024.
\newblock \href {https://aclanthology.org/2024.inlg-main.48/} {Leveraging large
  language models for building interpretable rule-based data-to-text systems}.
\newblock In \emph{Proceedings of the 17th International Natural Language
  Generation Conference}, pages 622--630, Tokyo, Japan. Association for
  Computational Linguistics.

\bibitem[{White and Baldridge(2003)}]{white-baldridge-2003-adapting}
Michael White and Jason Baldridge. 2003.
\newblock \href {https://aclanthology.org/W03-2316/} {Adapting chart
  realization to {CCG}}.
\newblock In \emph{Proceedings of the 9th {E}uropean Workshop on Natural
  Language Generation ({ENLG}-2003) at {EACL} 2003}, Budapest, Hungary.
  Association for Computational Linguistics.

\bibitem[{Wolf et~al.(2020)Wolf, Debut, Sanh, Chaumond, Delangue, Moi, Cistac,
  Rault, Louf, Funtowicz, Davison, Shleifer, von Platen, Ma, Jernite, Plu, Xu,
  Le~Scao, Gugger, Drame, Lhoest, and Rush}]{wolf-etal-2020-transformers}
Thomas Wolf, Lysandre Debut, Victor Sanh, Julien Chaumond, Clement Delangue,
  Anthony Moi, Pierric Cistac, Tim Rault, Remi Louf, Morgan Funtowicz, Joe
  Davison, Sam Shleifer, Patrick von Platen, Clara Ma, Yacine Jernite, Julien
  Plu, Canwen Xu, Teven Le~Scao, Sylvain Gugger, and 3 others. 2020.
\newblock \href {https://doi.org/10.18653/v1/2020.emnlp-demos.6} {Transformers:
  State-of-the-art natural language processing}.
\newblock In \emph{Proceedings of the 2020 Conference on Empirical Methods in
  Natural Language Processing: System Demonstrations}, pages 38--45, Online.
  Association for Computational Linguistics.

\bibitem[{Wyrwi\'{n}ski and Krawiec(2024)}]{wyrwinski}
Piotr Wyrwi\'{n}ski and Krzysztof Krawiec. 2024.
\newblock \href {https://doi.org/10.1145/3638530.3654277} {Guiding genetic
  programming with graph neural networks}.
\newblock In \emph{Proceedings of the Genetic and Evolutionary Computation
  Conference Companion}, GECCO '24 Companion, page 551–554, New York, NY,
  USA. Association for Computing Machinery.

\bibitem[{Yang et~al.(2024)Yang, Yang, Zhang, Hui, Zheng, Yu, Li, Liu, Huang,
  Wei, Lin, Yang, Tu, Zhang, Yang, Yang, Zhou, Lin, Dang, Lu, Bao, Yang, Yu,
  Li, Xue, Zhang, Zhu, Men, Lin, Li, Tang, Xia, Ren, Ren, Fan, Su, Zhang, Wan,
  Liu, Cui, Zhang, and Qiu}]{qwen2024qwen25}
An~Yang, Baosong Yang, Beichen Zhang, Binyuan Hui, Bo~Zheng, Bowen Yu,
  Chengyuan Li, Dayiheng Liu, Fei Huang, Haoran Wei, Huan Lin, Jian Yang,
  Jianhong Tu, Jianwei Zhang, Jianxin Yang, Jiaxi Yang, Jingren Zhou, Junyang
  Lin, Kai Dang, and 23 others. 2024.
\newblock \href {https://arxiv.org/abs/2412.15115} {Qwen2.5 technical report}.
\newblock \emph{arXiv preprint arXiv:2412.15115}.

\bibitem[{Zhang et~al.(2020)Zhang, Kishore, Wu, Weinberger, and
  Artzi}]{bert-score}
Tianyi Zhang, Varsha Kishore, Felix Wu, Kilian~Q. Weinberger, and Yoav Artzi.
  2020.
\newblock \href {https://openreview.net/forum?id=SkeHuCVFDr} {{BERTScore}:
  Evaluating text generation with {BERT}}.
\newblock In \emph{International Conference on Learning Representations}.

\bibitem[{Zhang et~al.(2021)Zhang, Tiňo, Leonardis, and Tang}]{9521221}
Yu~Zhang, Peter Tiňo, Aleš Leonardis, and Ke~Tang. 2021.
\newblock \href {https://doi.org/10.1109/TETCI.2021.3100641} {A survey on
  neural network interpretability}.
\newblock \emph{IEEE Transactions on Emerging Topics in Computational
  Intelligence}, 5(5):726--742.

\bibitem[{Zheng et~al.(2023)Zheng, Chiang, Sheng, Zhuang, Wu, Zhuang, Lin, Li,
  Li, Xing, Zhang, Gonzalez, and Stoica}]{zheng2023judging}
Lianmin Zheng, Wei-Lin Chiang, Ying Sheng, Siyuan Zhuang, Zhanghao Wu, Yonghao
  Zhuang, Zi~Lin, Zhuohan Li, Dacheng Li, Eric~P. Xing, Hao Zhang, Joseph~E.
  Gonzalez, and Ion Stoica. 2023.
\newblock \href {https://arxiv.org/abs/2306.05685} {Judging llm-as-a-judge with
  mt-bench and chatbot arena}.
\newblock \emph{arXiv preprint arXiv:2306.05685}.

\end{thebibliography}

\appendix

\section{Prompts of LLM Agents}
\label{app:prompt}
The prompts used for Software Architect, Software Engineer, Evaluator, Code Analyst and Test Engineer can be found in Fig.~\ref{fig:promptag1}, \ref{fig:promptag2}, \ref{fig:promptag3}, \ref{fig:promptag4}, \ref{fig:promptag5}, respectively.

In Figure~\ref{fig:promptd2t}, we show the prompt used for the zero-shot prompted LLM baseline to generate triple verbalizations directly.

All prompts are templates, with placeholders containing the specific data instances denoted by ``\texttt{\{name\}}", i.e.~they follow the Python string formatting convention.

\begin{figure*}[tp]
\begin{lstlisting}
You are an experienced software architect specializing in rule-based Natural Language Generation (NLG) systems implemented in Python. Your task is to provide high-level design guidance. You do not write implementation code. Instead, you define the structure of the system by specifying functions and their responsibilities.

When given a task, respond with:

- A concise description of the overall architecture.

- A list of functions (or classes, if needed), each with:
   - A clear signature.
   - A short description of its purpose.
   - Expected inputs and outputs.
- Optionally, a sketch of how components interact (e.g. as a sequence or flowchart).
- Do not write any implementation code. Your focus is on the design and structure of the system.  

# Your task is as follows.

Write a rule-based NLG system in Python for data-to-text task. Specifically, write a NLGSystem class with a function `verbalize_set_of_triples(triples)` that converts a list of RDF triples into plain text. 
Each RDF triple is an object containing the following properties: `triple.subject`, `triple.predicate` and `triple.object`. 
The possible values of `triple.predicate` are: {possible_predicates}
	
Example:
```
    
    triple1 = RDFTriple(subject = "School of Business", predicate = "academic staff size", object = "737")
    triple2 = RDFTriple(subject = "School of Business", predicate = "birth country", object = "Denmark")
    triples = [triple1, triple2]
    nlg = NLGSystem()
    output = nlg.verbalize_set_of_triples(triples) 
    # output should be e.g. "Denmark's School of Business has an academic staff size of 737 people."
```
Note that the subject of all RDF triples will not always be the same, and the list of triples may be shorter or longer than in this example. In some inputs, the subject of one triple may be the object of another, and so on. Make sure that your code generalizes well to all these cases. The generated text should contain all the information expressed in the triples while being fluent.
	
# Previously, you came up with the following design.
```
{design}
```

# The implementation provided by software engineers passed {num_test} unit tests, but failed the following:
{errors}

# Please come up with a new design for the system. You can use the previous design as a starting point, but you are not required to do so. You can also change the function signatures and names if you want to. Nevertheless, the whole implementation of NLG system should be in a single NLGSystem class, so in fact you need to design a list of functions for this class. Remember to include `verbalize_set_of_triples(triples)` function in your design.
\end{lstlisting}
\caption{Prompt of the Software Architect}
\label{fig:promptag1}
\end{figure*}

\begin{figure*}[tp]
\begin{lstlisting}
You are a skilled software engineer with strong Python expertise, tasked with implementing rule-based Natural Language Generation (NLG) systems. You work from high-level designs provided by a software architect and are responsible for writing clean, modular code that adheres to the specified structure.

Respond with Python code only.

# The description of the task is the following.

Write a rule-based NLG system in Python for data-to-text task. Specifically, write a NLGSystem class with a function `verbalize_set_of_triples(triples)` that converts a list of RDF triples into plain text. 
Each RDF triple is an object containing the following properties: `triple.subject`, `triple.predicate` and `triple.object`. 
The possible values of `triple.predicate` are: {possible_predicates}
	
Example:
```
    
    triple1 = RDFTriple(subject = "School of Business", predicate = "academic staff size", object = "737")
    triple2 = RDFTriple(subject = "School of Business", predicate = "birth country", object = "Denmark")
    triples = [triple1, triple2]
    nlg = NLGSystem()
    output = nlg.verbalize_set_of_triples(triples) 
    # output should be e.g. "Denmark's School of Business has an academic staff size of 737 people."
```
Note that the subject of all RDF triples will not always be the same, and the list of triples may be shorter or longer than in this example. In some inputs, the subject of one triple may be the object of another, and so on. Make sure that your code generalizes well to all these cases. The generated text should contain all the information expressed in the triples while being fluent.
	

# The current implementation of the system is as follows:
```
{program}
```

# This implementation passed {num_test} unit tests, but failed the following:
{errors}

# The design proposed by software architect is as follows.
{idea}

# To fix (even if only partially) these errors, you should rewrite `{func_name}` function from your code. 
# You cannot modify other functions, do not repeat the implementation of NLGSystem class. Output only the code of the `{func_name}` function. 

\end{lstlisting}
\caption{Prompt of the Software Engineer}
\label{fig:promptag2}
\end{figure*}

\begin{figure*}[tp]
\begin{lstlisting}
You are a careful evaluator of NLG systems. Given a set of input RDF triples and an output of data-to-text system, you evalute wheter the output is a correct verbalization of the input.
The system output is correct if it all facts expressed in the input triples are verbalized and no additional or incorrect infomation is mentioned. The output should be fluent and not repetitive.
You must answer strictly with 'correct' or 'incorrect'.

Input: {sample.data}
System output: {output}

Is the system output correct?
\end{lstlisting}
\caption{Prompt of the Evaluator}
\label{fig:promptag3}
\end{figure*}

\begin{figure*}[tp]
\begin{lstlisting}
You are an intelligent code analysis agent tasked with evaluating the current state of a rule-based Natural Language Generation (NLG) system in Python. You receive input from three sources:
    Architect: A high-level design specification listing functions, their purposes, and expected inputs/outputs.
    Engineer: The actual Python code implementing these functions.
    Evaluator: The test results, including passed/failed unit tests, error messages, and observed vs. expected outputs.

Your job is to analyze these three sources and determine:
    Whether a specific function is incorrectly implemented and needs to be fixed.
    Or whether the architectural design is flawed and requires a rethinking of the design or function definitions.

When responding, follow this format:
    Diagnosis Summary:
        Clearly state whether the issue lies in the implementation, the design, or both.
        Specify the affected function(s).
    Reasoning:
        Justify your diagnosis using evidence from the code and test results.
        Refer to discrepancies between the architect's intent and the engineer's implementation.
        Consider if the function's purpose or interface was unclear or unrealistic.
    Recommendation:
        If the implementation is flawed, suggest how the engineer might fix it (e.g., logic correction, better input validation).
        If the design is flawed, propose a revised high-level design for the problematic function or module.
Focus on clarity, accuracy, and actionable guidance. Be rigorous but constructive---your goal is to improve the system collaboratively.

### Task description

Write a rule-based NLG system in Python for data-to-text task. Specifically, write a NLGSystem class with a function `verbalize_set_of_triples(triples)` that converts a list of RDF triples into plain text. 
Each RDF triple is an object containing the following properties: `triple.subject`, `triple.predicate` and `triple.object`. 
The possible values of `triple.predicate` are: {possible_predicates}
	
Example:
```
    triple1 = RDFTriple(subject = "School of Business", predicate = "academic staff size", object = "737")
    triple2 = RDFTriple(subject = "School of Business", predicate = "birth country", object = "Denmark")
    triples = [triple1, triple2]
    nlg = NLGSystem()
    output = nlg.verbalize_set_of_triples(triples) 
    # output should be e.g. "Denmark's School of Business has an academic staff size of 737 people."
```
Note that the subject of all RDF triples will not always be the same, and the list of triples may be shorter or longer than in this example. In some inputs, the subject of one triple may be the object of another, and so on. Make sure that your code generalizes well to all these cases. The generated text should contain all the information expressed in the triples while being fluent.
\end{lstlisting}
\caption{Prompt of the Code Analyst (Part 1/2, continued on the next page)}
\label{fig:promptag4}
\end{figure*}

\begin{figure*}[tp]
\begin{lstlisting}
### Design
{idea}

### Implementation
{program}

### Evaluation
This implementation passed {num_test} unit tests, but failed the following:
{errors}

### What to do to fix these errors? Should I change the system design? Or fix some function?
\end{lstlisting}
\caption{Prompt of the Code Analyst (Part 2/2, cont.)}
\label{fig:promptag4cont}
\end{figure*}

\begin{figure*}[tp]
\begin{lstlisting}
You are an expert data generator. Your task is to generate a dataset for data-to-text task. 

Your task is to generate a dataset for data-to-text task. More precisely, for converting RDF triples into plain text. Each example in the dataset should contain: input (a set of RDF triples) and output (verbalization). For instance:		

Input: [RDFTriple(subject='Pontiac Rageous', predicate='production start year', object='1997'), RDFTriple(subject='Pontiac Rageous', predicate='assembly', object='Michigan'), RDFTriple(subject='Pontiac Rageous', predicate='production end year', object='1997')]
Output: 'Pontiac Rageous was first made in Michigan in 1997 and was last produced in 1997.'
	
In the generated dataset, possible `predicate` values of RDF triple are: {predicates}.

Below you have an example of RDF triple for every predicate.
{examples}

You can use RDF triples from examples above, but it is expected that you will generate new triples to construct new examples for the dataset. Note that the input may contain a single triple or multiple triples.

Generate {examples_per_request} diverse examples, each containing: input (a set of RDF triples) and output (verbalization).
 
\end{lstlisting}
\caption{Prompt of the Test Engineer}
\label{fig:promptag5}
\end{figure*}

\begin{figure*}[tp]
\begin{lstlisting}
You are given the following list of RDF triples.
{triples}
Write a plain text description of this data. Output only the text of the description.
\end{lstlisting}
\caption{Prompt for the zero-shot prompted LLM direct data-to-text generation baseline.}
\label{fig:promptd2t}
\end{figure*}

\section{Prompts for LLM-as-a-Judge}
\label{app:prompt-eval}
The prompt used to assess grammaticality is provided in Fig.~\ref{fig:prompt-gram}.
The prompt used to assess ommisions is provided in Fig.~\ref{fig:prompt-om}.
The prompt used to assess additions is provided in Fig.~\ref{fig:prompt-add}.

All prompts are templates, with placeholders containing the specific data instances denoted by ``\texttt{\{name\}}", i.e.~they follow the Python string formatting convention.

\begin{figure*}[tp]
\begin{lstlisting}
You are an expert evaluator of data-to-text generation task.

Your task is to evaluate the output of a data-to-text task, for which the model was instructed to produce a verbalisation of a given set of RDF triples. 
    
    You should assess **the grammatical correctness** of the resulting text. Do not take any other factors into account. Do not make assumptions or consider external knowledge not present in the provided context. Identify only errors relating to the grammaticality of the text. Do not consider aspects such as fluency, omissions or hallucinations.

Respond with 1 for correct and 0 for incorrect.

System output: {output}

Assess the grammatical correctness of the output. Answer with a single number 1 (correct) or 0 (incorrect), without any other text.
\end{lstlisting}
\caption{Prompt for assesing the grammaticallity of the provided output.}
\label{fig:prompt-gram}
\end{figure*}

\begin{figure*}[tp]
\begin{lstlisting}
You are an expert evaluator of data-to-text generation task.

Your task is to evaluate the output of a data-to-text task, for which the model was instructed to produce a verbalisation of a given set of RDF triples. 
    
You should assess the **omissions** in the resulting text; in other words, you should check whether any of the input triples were not verbalised.  You can perform the task by iterating over the input triples and checking if it is present in the output. Do not take any other factors into account. Do not make assumptions or consider external knowledge not present in the provided context. Identify only errors relating to the fluency of the text. Do not consider aspects such as grammaticality, fluency or the addition of new facts (hallucinations).

Respond with 1 if any of the input triples is ommited and 0 if not.

Input triples: {sample}
System output: {output}

Assess the omissions of the input triples. Answer with a single number: 1 (omissions) to 0 (no omissions), without any other text.

\end{lstlisting}
\caption{Prompt for assesing presence of ommisions in the provided output.}
\label{fig:prompt-om}
\end{figure*}

\begin{figure*}[tp]
\begin{lstlisting}
You are an expert evaluator of data-to-text generation task.

Your task is to evaluate the output of a data-to-text task, for which the model was instructed to produce a verbalisation of a given set of RDF triples. 
    
You should assess the **addition of new facts** in the resulting text which were not present in the input. You can perform the task by carefully reading the text and checking if the facts mentioned are present in the input triples. Do not take any other factors into account. Do not make assumptions or consider external knowledge not present in the provided context. Identify only errors relating to the fluency of the text. Do not consider aspects such as grammaticality, fluency or the omissions of input triples.

Respond with 1 if the output contains facts not mentioned in the input and 0 if not.

Input triples: {sample}
System output: {output}

Assess the additions of new facts in the output. Answer with a single number: 1 (additions) or 0 (no additions), without any other text.
\end{lstlisting}
\caption{Prompt for assesing presence of additions in  the provided output.}
\label{fig:prompt-add}
\end{figure*}

\section{Hyperparameters of BART fine-tuning}
\label{app:trainingdetails}

We used the BART-base model\footnote{\url{https://huggingface.co/facebook/bart-base}} with the default architecture for conditional language modelling provided by the HuggingFace library~\cite{wolf-etal-2020-transformers}.
AdamW with learning rate $\eta = 2\cdot 10^{-5}$ and parameters $\beta=(0.9,0.997)$, $\epsilon = 10^{-9}$ was used as optimizer.\
Additionally, we applied  polynomial scheduler of $\eta$ with a warmup equal to 10\% of optimization steps.
The training was scheduled for 20 epochs with early stopping on validation loss (patience of 10 epochs).
We used batch size equal to 8 and label smoothing with $0.1$ smoothing factor.

\section{Pseudocode}
\label{app:code}
The pseudocode of the proposed approach is presented in Alg.~\ref{alg:training}.

\begin{algorithm*}
\caption{Training procedure of our NLG system}
\label{alg:training}
\begin{algorithmic}[1]
\State \textbf{Input:} $KG$ - knowledge graph
\State \textbf{Output:} Functional \texttt{NLGSystem} class
\State \textbf{LLM Agents:} $SA$ - Software Architect, $TE$ - Test Engineer, $SE$ - Software Engineer, $Eval$ - Evaluator, $CA$ - Code Analyst

\Procedure{TrainNLGSystem}{}
    \State predicates, example\_triplets $\gets$ get\_all\_predicates\_with\_examples($KG$)
    \State unit\_tests $\gets$ $TE$.GenerateUnitTests(predicates, example\_triplets)
    
    \State list\_of\_functions, design $\gets$ $SA$.GenerateDesign(predicates)
    \State program $\gets \emptyset$ 
    \For{func in  list\_of\_functions}
	    \State program[func] $\gets$  $SE$.ImplementFunction(func, design)
    \EndFor

    \While{time limit not exceeded}
        \State output $\gets$ $Eval$.Run(program, unit\_tests)
        \State test\_results $\gets$ $Eval$.EvaluateOutputs(output)

        \If{AllTestsPass(test\_results)}
            \State \Return program
        \Else
            \State decision, feedback $\gets$ $CA$.Analyze(design, program, test\_results)
            \If{decision == "redesign"}
		    \State list\_of\_functions, design $\gets$ $SA$.GenerateDesign(predicates, example\_triplets, feedback)
		    \State program $\gets \emptyset$ 
		    \For{func in  list\_of\_functions}
			    \State program[func] $\gets$   $SE$.ImplementFunction(func, design)
		    \EndFor
            \ElsIf{decision == "refactor"}
		    \For{func in  decision.func\_to\_refactor}
    			\State program[func] $\gets$ $SE$.ImplementFunction(func, design, feedback)
		    \EndFor
            \EndIf
        \EndIf
    \EndWhile
\EndProcedure

\end{algorithmic}
\end{algorithm*}

\section{Human annotators}
All of the annotators are aged between 20 and 40, hold at least a Master's degree in Computer Science, and have expertise in NLG systems. Four of the annotators were European and two were Indian. The annotators were not paid specifically for performing the annotations, but were hired by our institution.

\section{Comparison with previous rule-based algorithms}
\label{app:rule-base}
We provide the comparison with the most related approach~\cite{warczynski-etal-2024-leveraging-large}, which also uses LLM to construct templates for RDF-to-text generation. 
The results are presented in Table~\ref{tab:warczy}.
The approach uses reference texts during training and is not able to work on out-of-domain examples. 
The approach generates over 113 000 rules to handle different cases in RDF-to-triple generation. To handle the same dataset, our approach generates only 16 programs (one for each domain), providing better interpretability.

\begin{table*}[t]
\small
\centering
\begin{tabular}{lrr|rr|rr|rr|c}
\toprule
                       & \multicolumn{2}{c}{\bf BLEU}          & \multicolumn{2}{|c}{\bf METEOR}        & \multicolumn{2}{|c}{\bf BERTScore}     & \multicolumn{2}{|c|}{\bf BLEURT} &  \bf Inter-                \\
                       & \bf All             & \bf OOD             &\bf  All             & \bf OOD             & \bf All             & \bf OOD             & \bf All              & O\bf OD             & \bf pretability \\\midrule
             
\multicolumn{10}{l}{\it Rule-based approach presented in~\cite{warczynski-etal-2024-leveraging-large}}                                                                                                      \\\midrule
trained by Llama 3 70B & \bf 0.3284          & n/\ a          & 0.4781         & n/\ a          & 0.8822          & n/\ a          & -0.4139          & n/\ a         & \cmark                \\\midrule
\multicolumn{10}{l}{\it Our rule-based NLG}       \\\midrule
trained by Llama 3.3 70B          & 0.2858          & \bf 0.2858          & \bf 0.6578          & \bf 0.6606          & \bf 0.9179          & \bf 0.9187          &      \bf 0.0762           &  \bf 0.0618               & \cmark         \\
\bottomrule        
\end{tabular}
\caption{Results of evaluation on the WebNLG dataset. BLEU, METEOR, BERTScore and BLEURT metrics are reported for the entire test set (All) and for out-of-domain examples (OOD). }
\label{tab:warczy}
\end{table*}
\end{document}